\documentclass[preprint,11pt]{elsarticle}

\usepackage{amssymb}
\usepackage{algorithm}
\usepackage{algorithmic}
\usepackage{amsmath}
\usepackage{subfig}

\graphicspath{ {./figures/} }

\journal{Artificial Intelligence Journal}

\begin{document}
\begin{frontmatter}



\title{Efficient Open-world Reinforcement Learning via Knowledge Distillation and Autonomous Rule Discovery}


\author{Ekaterina Nikonova}
\author{Cheng Xue}
\author{Jochen Renz}

\affiliation{organization={School of Computing},
            addressline={The Australian National University}, 
            city={Canberra},
            postcode={2601}, 
            state={ACT},
            country={Australia}}

\begin{abstract}
Deep reinforcement learning suffers from catastrophic forgetting and sample inefficiency making it less applicable to the ever-changing real world. However, the ability to use previously learned knowledge is essential for AI agents to quickly adapt to novelties. Often, certain spatial information observed by the agent in the previous interactions can be leveraged to infer task-specific rules. Inferred rules can then help the agent to avoid potentially dangerous situations in the previously unseen states and guide the learning process increasing agent's novelty adaptation speed. In this work, we propose a general framework that is applicable to deep reinforcement learning agents. Our framework provides the agent with an autonomous way to discover the task-specific rules in the novel environments and self-supervise it's learning. We provide a rule-driven deep Q-learning agent (RDQ) as one possible implementation of that framework. We show that RDQ successfully extracts task-specific rules as it interacts with the world and uses them to drastically increase its learning efficiency. In our experiments, we show that the RDQ agent is significantly more resilient to the novelties than the baseline agents, and is able to detect and adapt to novel situations faster. 
\end{abstract}



\begin{keyword}
open world learning \sep novelty adaptation \sep reinforcement learning
\end{keyword}

\end{frontmatter}


\section{Introduction}
\label{sec:introduction}


\begin{figure*}[h]
\centering
\includegraphics[scale=0.18]{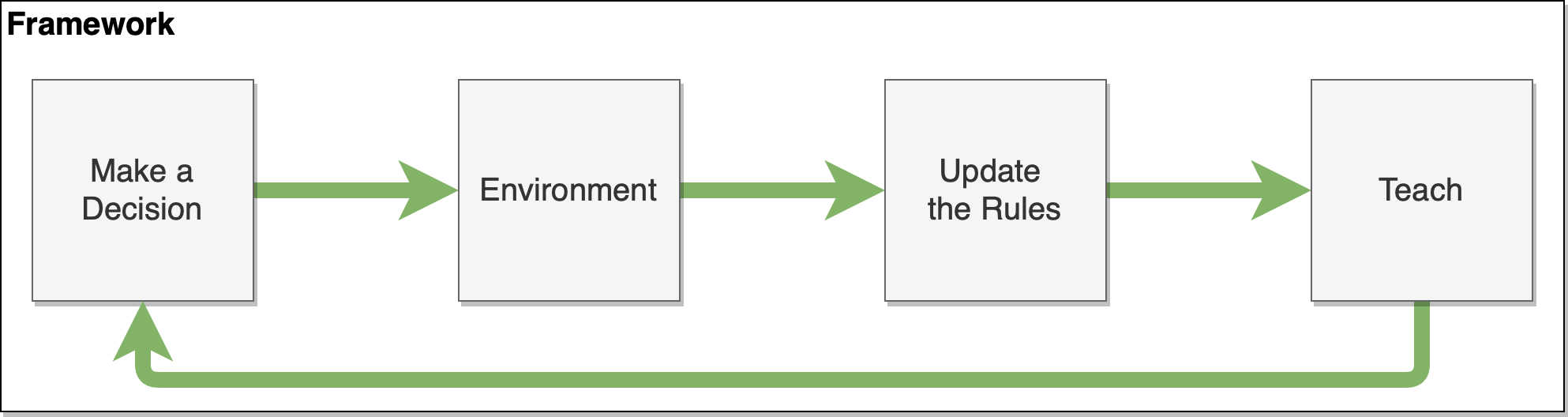}
\caption{The outline of the framework proposed in this work.}
\label{fig:framework}
\end{figure*}

One of the longstanding goals of Artificial Intelligence (AI) research is to efficiently generalize and reuse the experience obtained in one task in another. The ability to use previously learned knowledge is essential for AI agents to quickly adapt to new environments and to make learning more efficient in general. In recent years, deep reinforcement learning (DRL) agents have outperformed humans in various domains including Atari Games \cite{mnih2015humanlevel}, Go \cite{Silver2017MasteringTG}, and Dota 2 \cite{Berner2019Dota2W}. However, these methods require enormous computational power and time to learn how to perform well in a single environment let alone multiple. This limitation becomes especially evident when such agents are deployed in the real world. In the real world, there is a large number of possible novelties that agent needs to deal with, and often there is no preexisting data to train the agent for all of them. One of the potential ways to deal with novelties is to accumulate knowledge, generalize upon it and reuse it similarly to humans.

When interacting with a world, humans often generalize the learned experience in terms of relationships between them and objects. For example, if there is an unmovable obstacle in front of a human, one will try to avoid it indifferently if it is a wall, a hole in the ground, or someone's car. For the reinforcement learning agent, on the other hand, the sudden appearance of a previously unknown object can change its behavior dramatically. Often such behavior can be written in terms of relational rules, for example, "\textit{if there is an object in front of you do not go there}". Similar rules can be defined in terms of spatial relationships between the agent and other objects for example using qualitative spatial representation \cite{Clementini1997QualitativeRO, Cohn2008QualitativeSR, Chen2013ASO}.

Frequently, spatial rules are not unique to the particular environment and can be reused. Consider for example domains presented in Figure \ref{fig:learningrelations}. By interacting with the Super Mario Bros environment agent can learn that running into a Goomba (the enemy) will lead to its death and therefore should be avoided. By observing this scenario multiple times, the agent can generalize and infer the rule using spatial relationships between itself and Goomba. When a trained agent is brought to a new environment, such as for example, Frozen Lake, it can establish the direct mapping between Goomba and the hole and apply previously learned spatial rules.

When encountering novelty, such rules can then be used by the reinforcement learning agent to correct its policy and guide its behavior. For example, if the agent has previously died due to the collision with Goomba it has no need to run into Goomba again at the new level. This mimics how humans learn to interact with the environment, once human learns that Goomba is the enemy they will likely avoid it in all other experiences. We can use learned spatial rules to modify agent's policy and prevent it from doing actions that lead to undesirable consciences in the past. In particular, given a deterministic and discrete environment, we can use the spatial rules in conjunction with agents' policy to construct a new policy for each state. Continuing our example, if we have a spatial rule "\textit{if there is Goomba on the right of the agent do not go right}" we can use that rule to adjust the policy and assign probability zero to the action right. 

In this work, we propose a general framework that is inspired by these ideas. The proposed framework can be used with deep reinforcement learning agents to make learning more efficient, significantly improve the adaptation speed, and make the agent more resistant to the certain novelties. It consists of the four main components (Figure \ref{fig:framework}): a reinforcement learning agent, the environment which the agent interacts with, the rule-learning component, and knowledge distillation (from the teacher). We use the deep reinforcement learning agent to collect the experience necessary for learning and rule inference. We use inductive logic programming to infer rules that explain these observations. We focus only on explaining negative observations such as actions that lead to the immediate death of the agent. We then use inferred rules to guide the agent's learning process and distill them into the agent's policy.

For the experiments, we have provided an implementation of the proposed framework as part of a rule-driven deep Q-learning agent (RDQ). RDQ is a modified version of the vanilla deep Q-network \cite{mnih2015humanlevel} that autonomously learns the rules and self-supervises its learning. We test RDQ in three different domains against other state-of-the-art reinforcement learning algorithms. In our experiments we show that RDQ is significantly more sample efficient and resilient to the novelties, making overall training and adaptation to the novel scenarios drastically quicker.

\section{Related Work}
Open-world novelty accommodation has been an active area of research in recent years \cite{Goel2022RAPidLearnAF, Langley2020OpenWorldLF, Muhammad2021ANA}. In this work, we focus on the problem of novelty accommodation in reinforcement learning agents. 

\paragraph{\textbf{Reinforcement Learning}} In Reinforcement Learning (RL) several ideas were presented on the adaptation to the continuously evolving and non-stationary environments \cite{Khetarpal2020TowardsCR, Padakandla2019ReinforcementLA}. However, in this work, we focus on the adaptation to sudden changes. One of the possible approaches to novelty accommodation is the generalization and efficient use of the previous experience. RAPid-Learn, for example, is a hybrid planner and learner method that utilizes domain knowledge to perform knowledge-guided-exploration and adapt to novel scenarios \cite{Goel2022RAPidLearnAF}. In comparison to their work, RDQ agent does not require a planner and can be applied directly to model-free methods such as deep Q-learning. In general, in reinforcement learning there are several types of knowledge that can be transferred by the agent from one domain to another \cite{taylortransfer} including policy, value function, task model, options, macro-actions, skills, or observations \cite{Florensa2017ReverseCG, Silva2018ObjectOrientedCG, Shao2019StarCraftMW, Yang1996ProgressiveLA, Vezhnevets2016StrategicAW, Tessler2017ADH}. To deal with novelties we focus on reusing and generalizing the observations. One approach is to extract the rules using inductive logic programming from the observed data \cite{Xu2021InterpretableMH, Kimura2021NeuroSymbolicRL} or expert demonstrations \cite{Payani2020IncorporatingRB}. Our approach builds upon these previous works and uses inductive logic programming to extract relational rules. However, we focus only on explaining the negative experiences and use them to aid learning in novel scenarios rather than making a policy more interpretable. In our approach, only the inferred rules are explainable whether the policy is approximated using a neural network.

\paragraph{\textbf{Inductive Logic Programming}}
Inductive logic programming (ILP) \cite{Muggleton1994InductiveLP} is a form of machine learning which given a set of examples and a background knowledge induces the hypothesis which generalizes those examples. An ILP system represents a hypothesis as a set of logical rules that can be used by the agent directly. While learning rules to explain an agent's policy entirely has been tried before \cite{Glanois2021ASO}, we propose to focus only on the negative observations and let the reinforcement learning algorithm learn the rest. Similar ideas has been explored in ILP such as Popper \cite{Cropper2020LearningPB}. Popper is an ILP system that combines answer set programming and Prolog. Popper takes in background knowledge, negative examples, and bias to induce the explanation for the provided examples. Popper, similar to our work, focuses on learning from the failures. In their work they used it as constrains to prune the hypothesis space which in turn improved learning performance. We use Popper to extract rules that would explain negative observations. In comparison to their work, we use Popper in combination with reinforcement learning to abstract observations in the more general rules to aid adaptation.

\paragraph{\textbf{Safe Reinforcement Learning}}
Once extracted, the rules can be used together with the ideas presented in the safe reinforcement learning research. Thus for example we can use rules extracted from the negative examples to increase the safety of the reinforcement learning agent and decrease its search space. In general, incorporating safety into reinforcement learning to prevent the agent from doing harmful actions has been an active research topic in recent years \cite{Garca2015ACS, Hans2008SafeEF} and is known as \textit{safe exploration} \cite{Amodei2016ConcretePI}. One proposed approach to correct the agent's behavior was to completely overwrite its action if it was seen to be unsafe by the human overseer \cite{Saunders2018TrialWE}. Another approach used pre-computed (from the safety specifications) temporal logical rules as a mechanism of "shielding" the agent from the actions that can endanger it \cite{Alshiekh2017SafeRL}. Our approach builds upon these ideas and takes them further by eliminating the need for human knowledge and using inferred rules instead. This provides the agent with the agile ability to change the rules dynamically as the environment changes.

\paragraph{\textbf{QSR in Reinforcement Learning}}
To extract symbolic relational rules from the observations, we use qualitative spatial representation (QSR) \cite{Clementini1997QualitativeRO, Cohn2008QualitativeSR}. QSR provides a scalable and universal approach to encoding relational information observed by the agent. Previously, QSR has been used together with deep reinforcement learning agents and it has been shown to perform better than traditional reinforcement learning agents \cite{QSRRL}. We expand those ideas further and instead of directly using QSR state representation by the agent, we use it to learn the rules. We note that QSR is used only to learn the rules, and our framework allows reinforcement learning agents to use any type of representation as we will show in the experiments. There are many types of QSR one can use \cite{Chen2013ASO}, in this work we use cone-shaped directional representation \cite{Renz2004QualitativeDC} and a qualitative distance representation \cite{Frank1992QualitativeSR}.

\paragraph{\textbf{Knowledge distillation and teacher-student architecture}}
Distilling the knowledge from a complex model to a much simpler model is a technique known as distillation \cite{Bucila2006ModelC, Hinton2015DistillingTK}. Knowledge distillation (KD) compresses the knowledge from a big and complex model (teacher) to a smaller model (student). KD has been previously used in multi-agent reinforcement learning agents to compress the knowledge of several agents into a single model \cite{Gao2021KnowRUKR, Omidshafiei2018LearningTT}. In deep reinforcement learning, Kullback–Leibler divergence (KL) has proven to be one of the most effective techniques for distillation \cite{Rusu2015PolicyD} and has been used in autonomous driving \cite{Huang2021EfficientDR} and Atari Games \cite{Sun2019RealtimePD}. We build upon those ideas and use distillation to bring the agent's (student) policy to a constructed policy (teacher). In comparison to the work of \cite{Huang2021EfficientDR}, our framework does not require an expert's demonstrations and directly uses the agent's experience to construct a "teacher" policy from the inferred rules. 

\section{Deep Q-network}
In this work we consider a deterministic Markov Decision Process (MDP) $M = (S,A,T,r,\gamma)$, where $S$ is the state space, $A$ is the action space, $T:S \times A \rightarrow S$ the transition function, $r:S \times A \rightarrow r$ is a reward function, and $y \in [0, 1)$ the discount factor. 

A \textit{policy} $\pi : S \rightarrow A$ determines which action to take in each state. Typically, the goal of reinforcement learning is to find a policy that maximizes the expected discounted reward and is therefore considered to be \textit{optimal}. 

The \textit{Q-function} $Q^{\pi}(s,a) = \mathbb{E}^{\pi}[\sum^{\infty}_{t=0} \gamma^{t}r_{t}|s_{0}=s, a_{0} = a] $ measures the performance of the agent assuming it starts in a state $s$, takes action $a$ and follows the policy $\pi$ afterwards. 

The \textit{Value-function} $V^{\pi}(s) = \mathbb{E}^{a \sim \pi(s)}[Q^{\pi}(s,a)] $ measures the overall value of the state. Same as with policy, those functions can be optimal: $Q^{*}(s,a) = max_{\pi} Q^{\pi}(s,a) $ and $V^{*}(s) = max_{\pi} V^{\pi}(s)$. 
Finally, the \textit{optimal policy} can be retrieved from $Q^{*}$ as follows:  $\pi^{*}(s) = argmax_{a} Q^{*}(s,a) $.

In deep reinforcement learning, $Q$-function can be approximated using a nonlinear function approximator such as a neural network $Q(s,a,\theta_i)$, where $\theta_i$ are the weights of the Q-network at the i-th iteration. However, when using a nonlinear function approximator together with reinforcement learning, it can become unstable or even diverge due to the following problems: a) the correlation in the sequence of observations, b) correlations between Q values and target values $ r_t + \gamma max_{a}Q(s_t,a) $ and c) having a policy that is extremely sensitive to changes of Q value. 

A deep Q-network (DQN) addresses the first problem by using \textit{experience replay}. Experience replay is implemented by storing and later randomly sampling the observations experienced by the agent. This technique removes the correlation between the sequences of the observations by randomizing the collected data. We define the experience as $e_t=(s_t,a_t,r_{t+1},s_{t+1})$, and experience set as $M=\{e_1,…,e_t\}$.

In order to address the second problem, the notion of target network was introduced which is then used to calculate the loss: $L_i(\theta_i) = \mathbb{E}_{(s,a,r,s^{'})\sim U(M)} [ (r+\gamma max_{a^{'}}Q(s^{'},a^{'},\theta_i^{-}) -Q(s,a,\theta_i ))^2 ]$, where $i$ is the iteration, $\gamma$ is a discount factor, $\theta_i$ are weights of so-called online Q-network and $\theta_i^{-}$ are weights of a target network or so-called offline Q-network. The target network is called offline since its weights are only updated every $C$ steps with a copy of online network weights, while the online network is updating every iteration $i$.

\begin{figure}[h]
\centering
\includegraphics[scale=0.2]{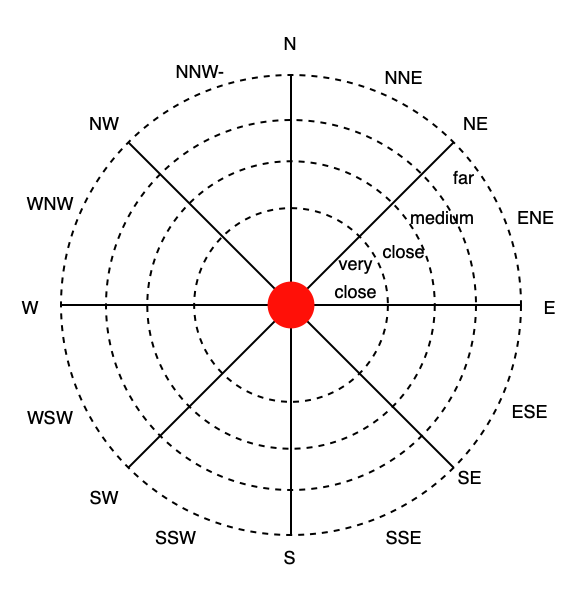}
\caption{Qualitative directional and distance representation used for the rules. }
\label{fig:qsr}
\end{figure}

\section{Qualitative Spatial Representation}

Often, to make a decision one needs to be aware of the type and nature of the surrounding objects. For example, if the agent detects a hole in the road in front of itself, it should avoid it. Such spatial information can be encoded using symbolic language. In this work, we construct a symbolic representation of the states using the extracted qualitative spatial relationships (QSR) between the objects and the agent. While there are many possible ways to encode such knowledge, we focus only on the direction and distance between the objects. As our representation languages, we use cone-shaped directional representation \cite{Renz2004QualitativeDC} and a qualitative distance representation \cite{Frank1992QualitativeSR}. 

Figure \ref{fig:qsr} shows an example of a combination of directional and distance representations. Here the red dot in the center represents an agent. We restrict the number of possible relationships, by only focusing on those that fall into a small square observation area around the agent (Figure \ref{fig:learningrelations}). In our experiments, we show that such representation is sufficient to make the learning process drastically more efficient. We use this representation to infer rules and construct a relational representation of the state.

\section{Symbolic Rules}
\label{sec:ruledriven}
\begin{figure*}
\centering
\includegraphics[width=.35\linewidth]{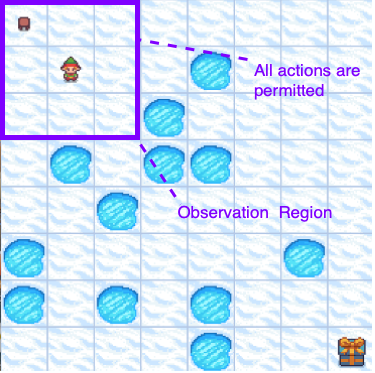}
\includegraphics[width=.36\linewidth]{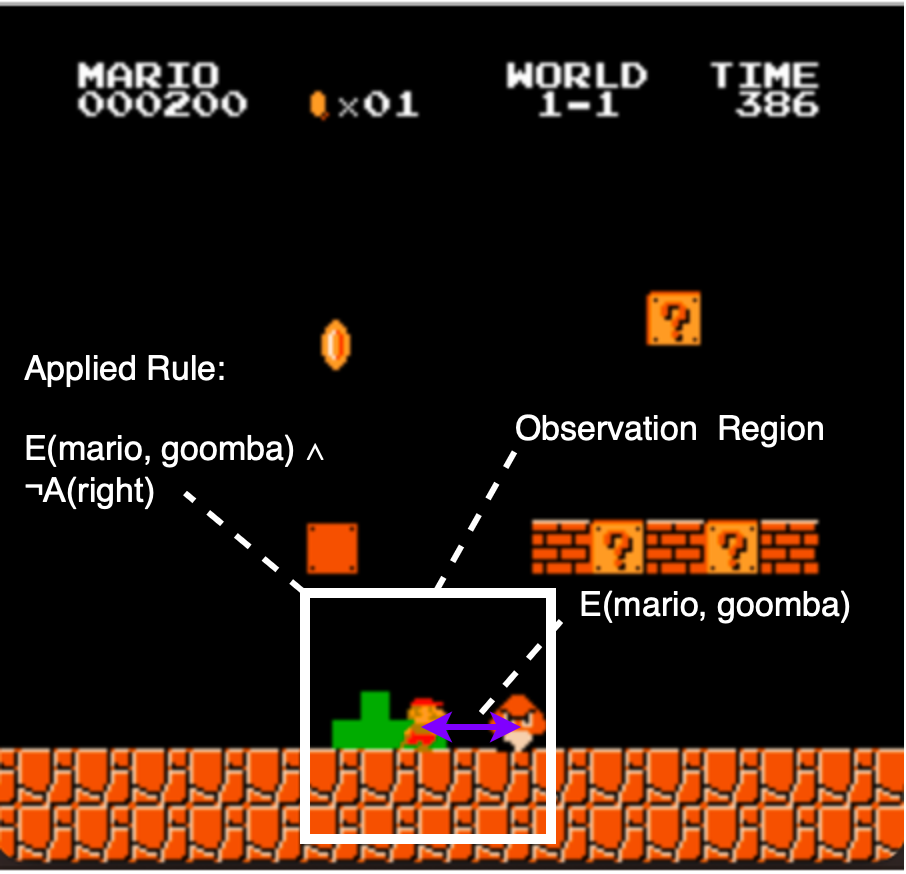}
\caption{Symbolic representation of FrozenLake and Super Mario Bros. Here square area around the agent demonstrates its observation field. Any object that is within it is assigned a QSR relationship and tested against all rules.}
\label{fig:learningrelations}
\end{figure*}

When humans encounter a new task, they tend to reuse previous knowledge rather than relearn everything from the scratch. Consider for example a scenario where a human drives a car and some object suddenly appears in front of the car. No matter whether it is an animal, a human, or some other unidentified object a human will very likely hit the brakes and try to avoid the collision. On contrary, the behavior of the reinforcement learning agent in this scenario is highly unpredictable. In this section, we propose to moderate the behavior of the agent in such novel scenarios by using symbolic rules. We note that rules such as "\textit{if there is an object in front of you, try to avoid it}" are rather universal and can be reused in a large number of domains and tasks.

Consider for example Figure \ref{fig:learningrelations} demonstrating spatial representation of the two domains used in this work. In Frozenlake, once the agent learns that it should avoid the holes, it should avoid it in all other levels as well independent if it has seen such state before or not.

We hypothesize, that by preventing the agent from performing unsafe actions we reduce the size of the state and action space that the agent should explore, thus increasing the performance. In this work we focus on avoiding actions that would lead to immediate failure, however, the proposed method can be adjusted to also prevent actions that would eventually lead to failure after some number of steps using for example a model-based approach \cite{Thomas2022SafeRL}.

\begin{figure*}[h]
\centering
\includegraphics[scale=0.12]{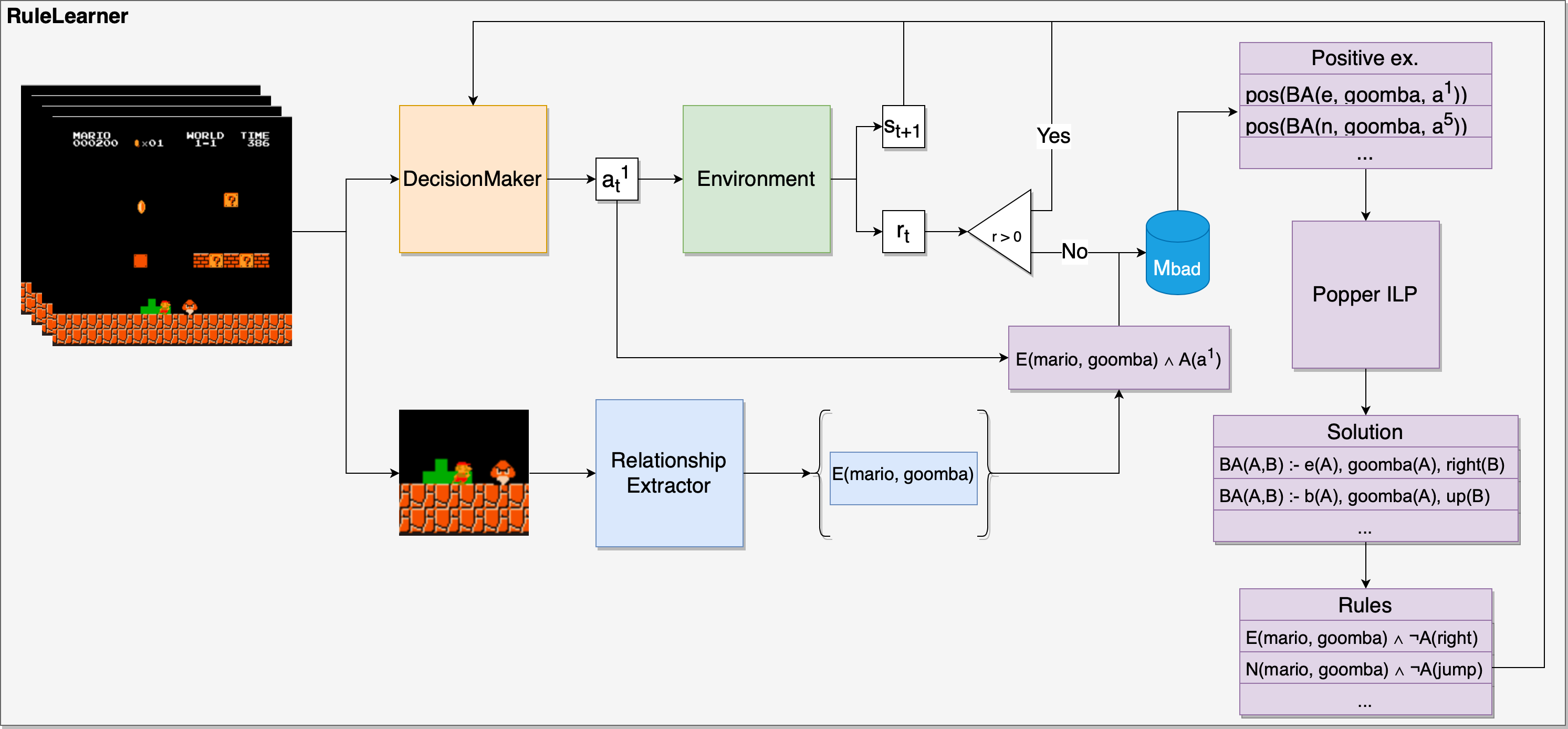}
\caption{Self-supervised rule-learning process. A reinforcement learning agent interacts with the environment and collects the experience needed to infer rules. Once sufficient experience is collected, an ILP is used to find an explanation for the negative experiences. That explanation is then fed back to the agent to teach and guide.}
\label{fig:rulelearner}
\end{figure*}

\paragraph{\textbf{Rule Definition}} We focus on the relationships between the objects to determine if the action is safe or not in the given state. We define a \textit{rule} to be a conjunction of n-ary relationships between objects and action which if satisfied would compromise safety: $(r_{1}(o_{1}, ... o_{n}) \wedge ... \wedge r_{m}(o_{1},...,o_{n}) \wedge \lnot action(a) )$, where $r_{i}$ is a QSR relationship, $o_{j}$ is an object and $action(a)$ is the action. Continuing our example, we can define a rule to prevent the collision with the nearby objects as: $close(agent, o) \wedge N(agent, o) \wedge \lnot action(up)$). In general, each domain would have a collection of such rules. Coming back to Figure \ref{fig:learningrelations}, for Frozen Lake, the agent would need to learn four rules: if the hole is in either north, east, south, or west direction and actions up, right, down, and left are unsafe.

\section{Self-supervised Rule Learning}

In theory, the rules of the game can be discovered automatically by the agent while interacting with the environment. Given a deterministic environment, confidence in such rules would increase as the agent collects more experience. For example, consider a scenario in Figure 3. The agent can observe that every time Mario moves right, it collides with the Goomba and receives a negative reward. By collecting enough samples of this scenario, the rule "\textit{if goomba is on the right of Mario, don't go right}" can be inferred. 

\begin{figure*}[h]
\centering
\includegraphics[scale=0.135]{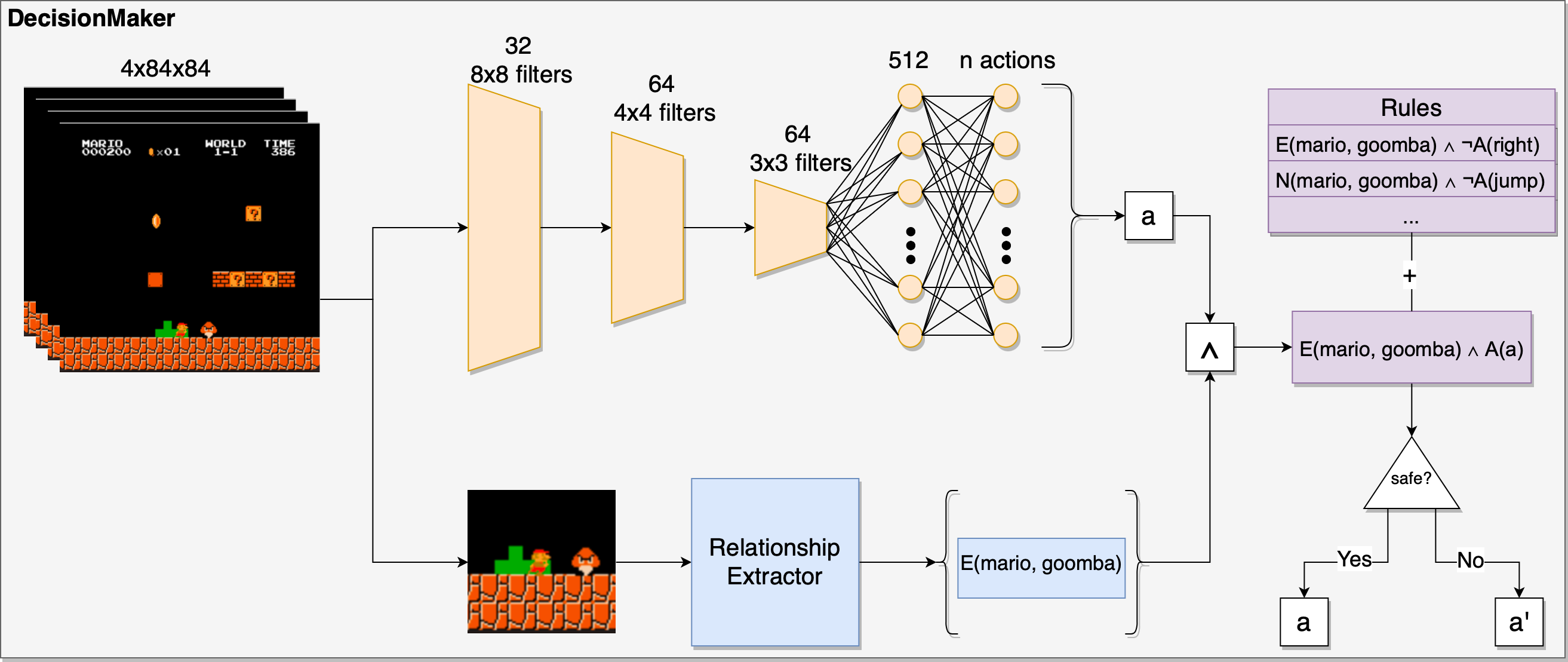}
\caption{A decision-making process inside RDQ agent. Here we use deep Q-learning to select the action and validate its safety according to the inferred rules. If the action is considered to be unsafe, such action is overwritten.}
\label{fig:decisionmaker}
\end{figure*}

In this work, we use inductive logic programming to infer the rules that would explain all negative observations, and use them to guide the agent in learning. The whole process of learning the rules is shown in Figure \ref{fig:rulelearner}.

While learning, the agent stores all $(s^{qsr}_{t}, a_{t})$ pairs that lead to a negative reward. We then convert observations to the positive examples and use inductive logic programming (Popper ILP) to infer the rules that would explain them. Finally we convert logical rules to the dictionary like structure inside (Figure \ref{fig:rulelearner} and Figure \ref{fig:learnedrule}) that it can be easily queried by the agent.

By doing so, we compress a potentially large number of observations to a relatively small number of rules for the agent to follow. To infer the rules from failures we use Popper - an ILP system that combines answer set programming and Prolog \cite{Cropper2020LearningPB}. The Popper takes in background knowledge, negative examples, and bias to induce the explanation for the provided examples. The rules are updated after a fixed number of steps until the agent's total reward for the episode is greater than preset threshold.

We theorize that once the agent encounters a novel environment, that is different enough from the previous one, it will cause a noticeable drop in performance. In reinforcement learning, a drop can be measured using the total reward per episode. If the total reward drops below a certain threshold, the algorithm would classify it as novelty and the agent will start learning rules again. By continuously allowing the agent to update its beliefs if they are no longer valid, we can ensure that the agent could adjust to the novelty.

\section{Decision-making Under The Rules}
Once the rules are inferred they need to be incorporated into the agent's learning process. In this section, we will look at using rules for guidance.

\begin{algorithm}
\caption{Safe $\epsilon$-greedy}
\label{alg: eps-greedy}
\textbf{Input}: $Q$, $s_{t}$ $s^{qsr}_{t}$ \\
\textbf{Output}: $a$ \\
\begin{algorithmic}[1] 
\STATE n $\sim$ $\mathcal{U}_{[0, 1]}$
\IF{n \textless $\epsilon$}
    \STATE $a_{t}$ $\gets$ \textit{selectRandomSafeAction}($s^{qsr}_{t}$)
\ELSE
    \STATE $a_{t}$ $\gets$ $argmax_{a} Q(s_{t}, a)$
    \IF{not \textit{isActionSafe}($s^{qsr}_{t}$, $a_{t}$)}
        \STATE $a_{t}$ $\gets$ \textit{selectRandomSafeAction}($s^{qsr}_{t}$)
    \ENDIF
\ENDIF

\RETURN $a_{t}$
\end{algorithmic}
\end{algorithm}

\paragraph{\textbf{Is the action safe?}} We call an action $a_{t}$ to be \textit{safe} in state $s_{t}$, if performing that action would not violate any known rule. To validate that action is safe, we extract symbolic relationships from the state $s_{t}$ as $s^{qsr}_{t} = \{ r_{1}(o_{1}, ... o_{n}),...,r_{m}(o_{1},...,o_{n}) \}$, where $r_{i}$ is a QSR relationship and $o_{j}$ is an object in the state $s_{t}$. We then conjugate it with the symbolic representation $a^{qsr}_{t}$ of action $a_{t}$ (i.e. $action(a_{t})$). Given that, we define 

$isActionSafe(s^{qsr}_{t}, a_{t})$ =

\begin{equation}
 \label{eq:isActionSafe}
  \begin{cases}
 1,& \text{if } (s^{qsr}_{t} \wedge \lnot a^{qsr}_{t})\cap rules = \emptyset \\
 0,& \text{otherwise}
 \end{cases}
\end{equation}

Consider Super Mario Bros on Figure \ref{fig:learningrelations}. Here the rules would prevent the action that would result in a collision with a Goomba. In this example, the Goomba is immediately in the right of the agent (i.e. $s^{qsr} = (close(agent, Goomba) \wedge E(agent, Goomba))$). If the RDQ predicts action "right" the rule would be violated and the algorithm would select a random safe action instead.

\paragraph{\textbf{Random safe action}} Given a state $s^{qsr}_t$, action space $A$ and safe actions $A_{safe}=\{a^{i}_{t} | isActionSafe(s^{qsr}_{t}, a^{i}_{t}) = 1, a^{i}_{t} \in A \}$ we can select a random safe action as:

$selectRandomSafeAction(s^{qsr}_t) =$
\begin{equation}
 \label{eq:selectRandomSafeAction}
\begin{cases}
a \in A_{safe},& \text{if } A_{safe} \neq \emptyset \\
a \in A,& \text{otherwise}
\end{cases}
\end{equation}
In Equation \ref{eq:selectRandomSafeAction}, action is sampled from $A_{safe}$ or $A$ uniformly.

\paragraph{\textbf{Safe $\epsilon$-greedy}}
We propose a method to inject the inferred rules as part of the modified, safer version of the $\epsilon$-greedy algorithm. \textit{Safe $\epsilon$-greedy} algorithm directly prevents the agent from performing unsafe actions by completely overriding its decision.

Recall the $\epsilon$-greedy policy \cite{suttonrlintro}:
\begin{equation}
 \label{eq:epsilongreedy}
\pi(s_t) = 
\begin{cases}
a \in $A$,& \text{if } n \textless \epsilon, n \in \mathcal{U}_{[0, 1]} \\
argmax_{a} Q(s_{t},a),& \text{otherwise}
\end{cases}
\end{equation}

\begin{figure*}[h]
\centering
\includegraphics[scale=0.12]{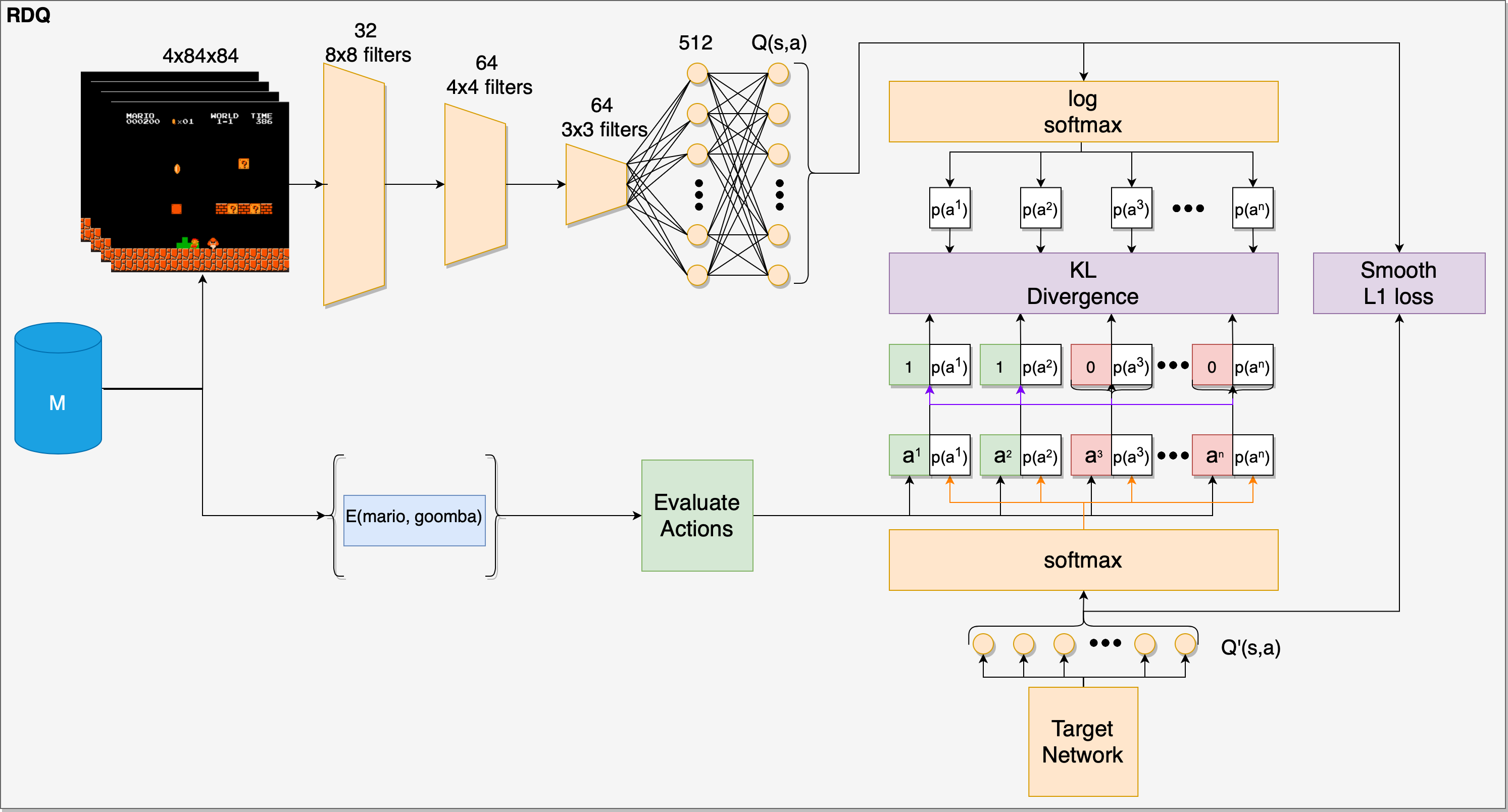}
\caption{Optimization flow in RDQ agent. Here we compute smooth L1 loss between target and online networks and additionally compute KL divergence between predicted (student) and constructed (teacher) policies. The teacher policy is constructed by modifying the predicted target policy using inferred rules.}
\label{fig:teacherstudent}
\end{figure*}

By definition, the epsilon-greedy policy (Equation \ref{eq:epsilongreedy}) selects a random action during the exploration phase, which, without any safety check, could result in the agent damaging itself or others. Instead, we propose to select a random safe action.

Algorithm \ref{alg: eps-greedy} demonstrates how the inferred rules can then be embedded into the $\epsilon$-greedy policy resulting in its safer version. Here instead of selecting a random action, the algorithm selects a random \textit{safe} action. In addition, when an action is selected using a learned $Q$-function, the algorithm will check whether this action is safe, and if not, it will overwrite such action. Ultimately, as the result of the safe $\epsilon$-greedy, many unsafe actions will not be explored thus reducing the size of the search space. We note that safe action will be added to the experience replay $M$ to aid further training.

\section{Rule-driven Deep Q-learning (RDQ)}

While preventing the agent from performing an unsafe action is an effective way to teach the agent only "good" actions, it does not prevent the agent from doing those actions once the safeguard is removed. In addition, one should adjust the agent's policy to teach it to avoid such actions in a more direct way. In particular, such actions should receive a very low probability of being performed. 

\paragraph{\textbf{Loss computation}} To address that, the RDQ agent uses inferred rules to evaluate each action in the action space and uses it to adjust the target policy, thus creating a new target (Figure \ref{fig:teacherstudent}). It then computes KL divergence between its policy $\pi^{student}(a_{t}|s_{t})$ and adjusted target policy $\pi^{teacher}(a_{t}|s_{t})$ as:

\begin{multline}
D^{KL}(\pi^{student}(a_{t}|s_{t}), \pi^{teacher}(a_{t}|s_{t}))
\label{klloss}
\end{multline}




Figure \ref{fig:teacherstudent} shows the overall process of optimization step in RDQ agent. A vanilla DQN computes the loss between online DQN Q-values and offline (target) DQN Q-values as:

\begin{multline}
    Q^{loss}_{i}(\theta_i) = \mathbb{E}_{(s_{t},a_{t},r_{t},s_{t+1})\sim U(M)} [ (r+\gamma max_{a^{'}}Q(s_{t+1},a_{t},\theta_i^{-})-Q(s_{t},a_{t},\theta_i ))^2 ]\label{loss}
\end{multline}

In \eqref{loss}, $i$ is the iteration, $\gamma$ is a discount factor, $\theta_i$ are weights of an online Q-network and $\theta_i^{-}$ are weights of a target network. We use standard DQN as the base for our algorithm, but in addition to computing error between online and target Q-values we compute KL divergence between the predicted and constructed policy:

\begin{equation*}
    L_i(\theta_i) = Q^{loss}_{i} + \lambda * D^{KL}_{i}
    \label{allloss}
\end{equation*}
Where $Q^{loss}_{i}$ is Equation \ref{loss}, $D^{KL}_{i}$ is Equation \ref{klloss} and $\lambda$ is a non-negative Lagrangian multiplier.

\paragraph{\textbf{Teacher policy construction}} In general we would like to completely avoid actions that violate inferred rules. Therefore, the probability of such actions should be zero in the teacher's policy. To construct the teacher policy, we take a predicted (by a target Q-network) target policy $\pi'(s_{t})=\{p(a_1), p(a_2), p(a_3)…, p(a_n)\}$ and adjust it by using rules. We define a set of unsafe actions as all actions except the ones that are safe: $A_{bad} = A \backslash A_{safe}$. We then define $p_{A_{bad}} = \sum_{a}^{A_{bad}} \pi'(a|s_{t})$ as a sum of probabilities of all bad actions. 

We can now define the constructed teacher policy as:
\begin{equation}
 \label{eq:teacherpolicy}
\pi^{teacher}(a|s_t) = 
\begin{cases}
\pi'(a|s_{t}) + \frac{p_{A_{bad}}}{|A_{bad}|}, &  \text{if } a \in A_{safe} \\
0, & \text{otherwise } 
\end{cases}
\end{equation}

In Equation \ref{eq:teacherpolicy} we use inferred rules to determine the safety of the action given QSR representation of the state $s^{qsr}_{t}$ and predicted target policy $\pi'(s_{t})$. If action is unsafe according to the rules, we assign a probability of zero to it. We accumulate all predicted probabilities of unsafe actions and then redistribute it equally among other "good" actions. Similarly to $Q^{loss}$ the examples are randomly sampled from $M$. The construction of teacher probability is happening at each optimization step and is used to compute $D^{KL}$ loss.

\section{Experimental Setting}

\begin{figure*}
\centering
\subfloat{\includegraphics[width=.3\linewidth]{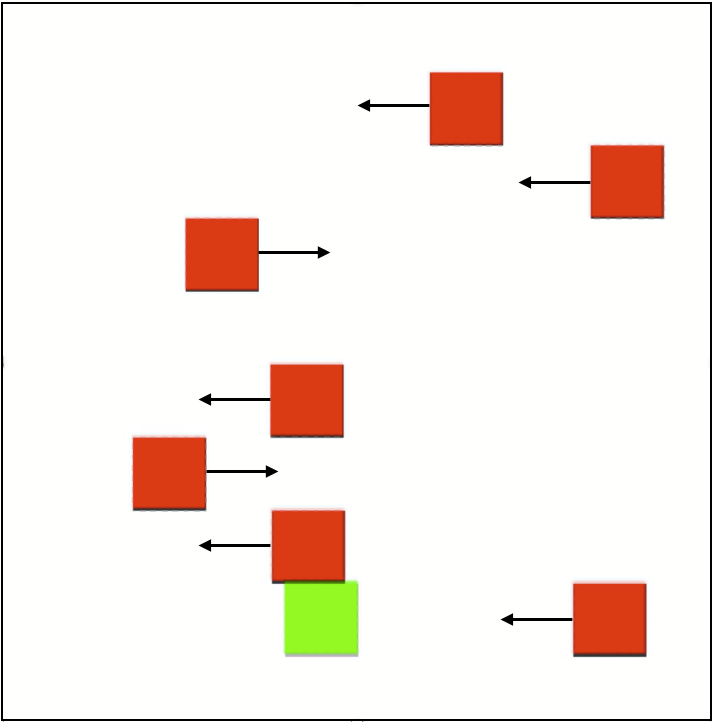}}
\subfloat{\includegraphics[width=.3\linewidth]{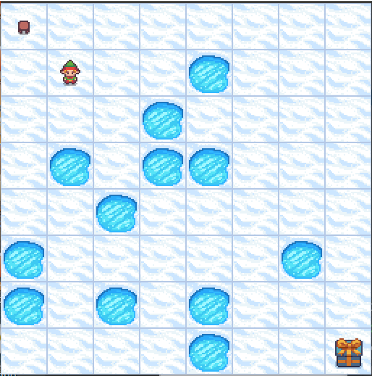}}
\subfloat{\includegraphics[width=.315\linewidth]{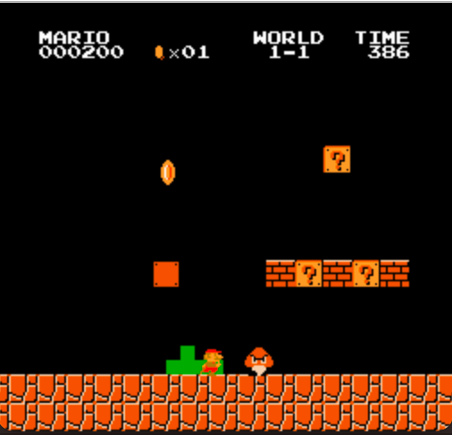}}
\caption{ Domains used in this work. From left to right: Crossroad, Frozenlake and Super Mario Bros.}
\label{fig:test-domains}
\end{figure*}

\paragraph{\textbf{Testing Domains}} 
To empirically evaluate the RDQ agent, we test it on three different domains: Crossroad, OpenAI FrozenLake, and OpenAI Super Mario Bros (Figure \ref{fig:test-domains}).

Crossroad is a discrete grid-like environment inspired by Atari Freeway, but has a bigger action space (5 vs 2 actions) and more importantly allows the injection of novelties. It has 7 cars moving horizontally at different speeds and directions and a player to control. The goal of the game is to cross all roads without being hit by a car (red boxes). 

FrozenLake is one of the environments provided by OpenAI Gym \cite{Brockman2016OpenAIG}. FrozenLake is also a discrete grid-like environment. The goal of the game is to navigate the player from its start state to the goal state without falling into the holes. We inject novelties into FrozenLake by generating random maps. Both of the domains have discrete action spaces. For FrozenLake we use RAM representation and for Crossroad we use a ground-truth representation that contains the positions of cars and the player. Both domains are available in open-source.

Super Mario Bros on the other hand is a more complex domain where the agent needs to complete the level while maximizing the number of collected coins and avoiding death. Similarly to FrozenLake, we use the OpenAI Gym version of Super Mario Bros. Contrary to the previous two domains, we use an image as our state representation. For all experiments with RDQ and baselines, we use the version with 7 possible actions ("simple movement") including ability to navigate to the left.

\paragraph{\textbf{Crossroad Novelties}}
For Crossroad the novelty can occur in either the velocity or direction of the cars. In total we have 9 different base novelties: Baseline, Super Slow speeds (all cars moving super slow requiring agent to wait or change its path), Super Fast speeds, Random speeds (speeds randomly drawn from the uniform distribution), Opposite speeds (same as baseline but cars move to the opposite sides), All cars moving left, All cars moving right, Shifted speeds (same as baseline but speeds are "shifted" by one position), and Reversed speeds (1st car now moves as last car in normal environment).
We add random noise to all 9 base novelties to generate 900 different levels, with 100 levels per novelty. We first train the RDQ agent on the baseline setting and then train the same model on the novelties. We note that we reset the model to its baseline trained state after each novelty.

\paragraph{\textbf{FrozenLake Novelties}}
For FrozenLake the novelty can occur either in the position of the holes or in the position of the goal and start states. In total we have two novelties: randomly shuffled holes positions and flipped along x-axis start and goal states. Similarly, as in Crossroad, we first train the model on the baseline (standard version of the game) and then generate 100 levels per novelty and retrain the same model on the new levels. We reset the model to the baseline model after each level.
\begin{figure*}
\centering
\subfloat{\includegraphics[width=.22\linewidth]{figures/domains/mario_11.png}}
\subfloat{\includegraphics[width=.228\linewidth]{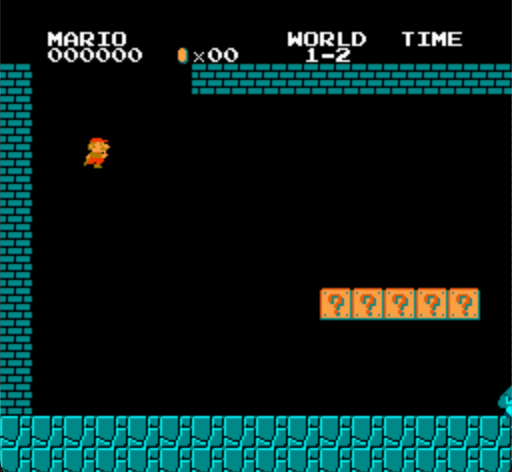}}
\subfloat{\includegraphics[width=.225\linewidth]{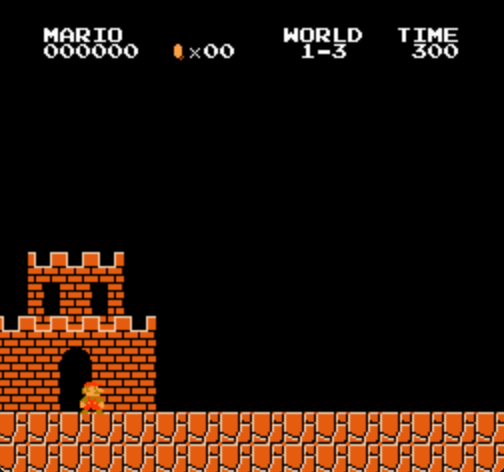}}
\subfloat{\includegraphics[width=.223\linewidth]{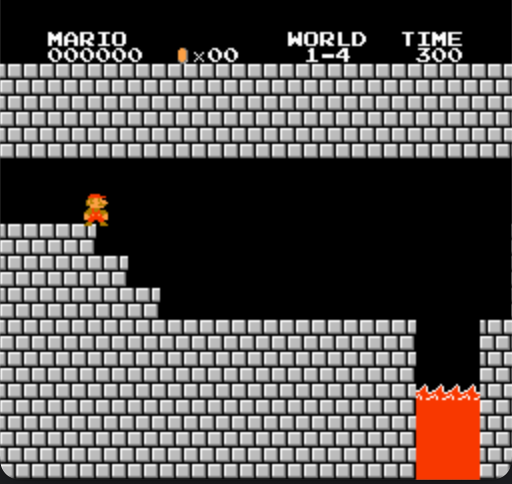}}
\caption{ Four levels in Super Mario Bros. From left to right 1-1, 1-2, 1-3, 1-4.}
\label{fig:mario-levels}
\end{figure*}

\paragraph{\textbf{Super Mario Bros Novelties}}
For Super Mario Bros, each level is novel by definition. For example, a new level can contain new enemies, has a different level layout, and new colors and objects. We test RDQ in four different levels 1-1, 1-2, 1-3, and 1-4 (Figure \ref{fig:mario-levels}). In our experiments, we first train the agent on level 1-1 and then train the same model on the different levels. Ultimately, Super Mario Bros is much harder than the previous two domains due to the larger number of possible novelties and complexity of those novelties.

\paragraph{\textbf{Agents and Settings}}
For Frozenlake and Crossroad, we use a ground truth representation (i.e. positions of the objects) as our state to simplify computational complexity. However, our method easily works with image representation as well and we show it in Super Mario Bros, where we use images as our states. In addition to the ground truth/image state, the agent receives the QSR representation of the observation field. Such QSR representation is a list of spatial relationships with objects that lie in the observation field of the agent. If there are no objects in the observation field, the agent receives the empty list (in such case any action is possible and purely depends on the agent's policy).

\begin{figure*}
\centering
\subfloat{\includegraphics[width=1.0\linewidth]{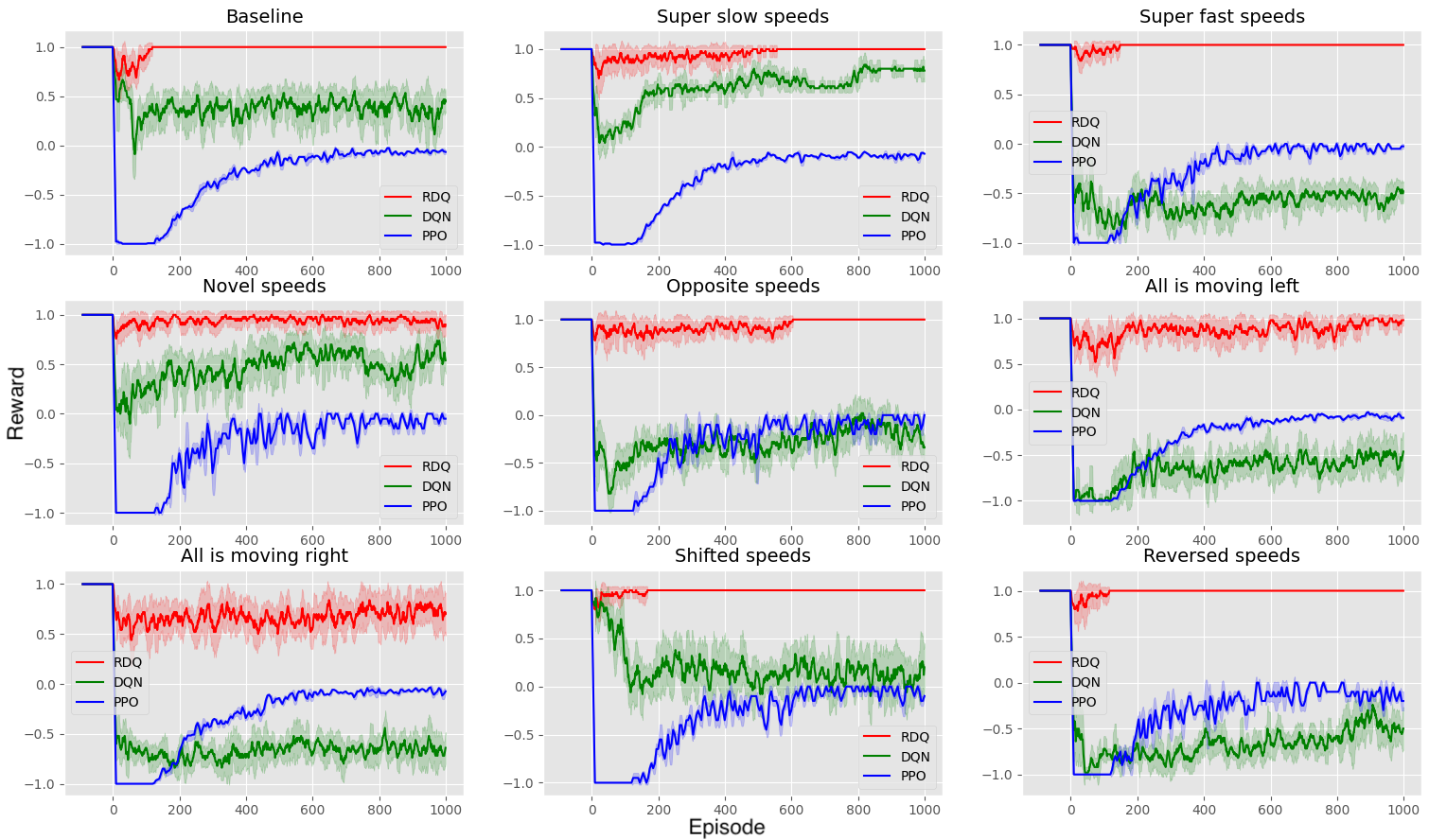}}
\caption{ Crossroad average results over 100 levels per each novelty. Here negative episodes represent pre-novelty performance. }
\label{fig:res-crossroad}
\end{figure*}

We compare the performance of the RDQ agent with two baselines: PPO \cite{Schulman2017ProximalPO} and DQN \cite{mnih2015humanlevel}. We chose DQN as one of the baselines since RDQ agent uses DQN internally and would allow us to show the pure benefit of using our framework. We note that the networks used in the RDQ and DQN agents are identical in their architecture and hyperparameters.
We chose PPO as our second baseline as it belongs to a different type of RL algorithm than DQN, and enables the comparison of RDQ to on-policy methods.

\paragraph{\textbf{QSR and Rules Setting}}
For all domains, we use identical QSR representation as described in the QSR section. We use QSR with a granularity of 64 and split the observation field into 64 regions. Each such region is assigned a unique symbol to enable rule-learning using Popper \cite{Cropper2020LearningPB}.

We store observations that lead to negative reward (immediate death) as $(s^{qsr}_{t}, a_{t})$ in the agent's memory $M_{bad}$. $M_{bad}$ is then used as positive examples for Popper to infer the rules. We filter out the "outlier" observations, i.e. $(s^{qsr}_{t}, a_{t})$ that were only observed less than some threshold. For all domains, the threshold is set to 10.

Once the agent encounters the novelty (i.e. its total episode reward drops below some preset threshold), the agent clears $M_{bad}$ as previous observations could potentially contradict the new ones and make the problem unsolvable.

\section{Results and Discussions}

\paragraph{\textbf{Crossroad}}
Figure \ref{fig:res-crossroad} shows the overall average results obtained by DQN, PPO and RDQ agents in 9 types of novelties (i.e. 900 levels with 100 levels per each novelty type). Before the novelties both agents (DQN and RDQ) were trained to their maximum performance on the base level, solving level completely. We note that in Crossroad, the RDQ agent shows resilience to the all types of novelties with only a slight performance drop and quickly recovers once the novelty is encountered. DQN and PPO on the other hand shows a dramatic performance drop in most of the novelties and does not recover in the limited number of episodes (1000 episodes).

\begin{figure*}
\centering
\subfloat{\includegraphics[width=.49\linewidth]{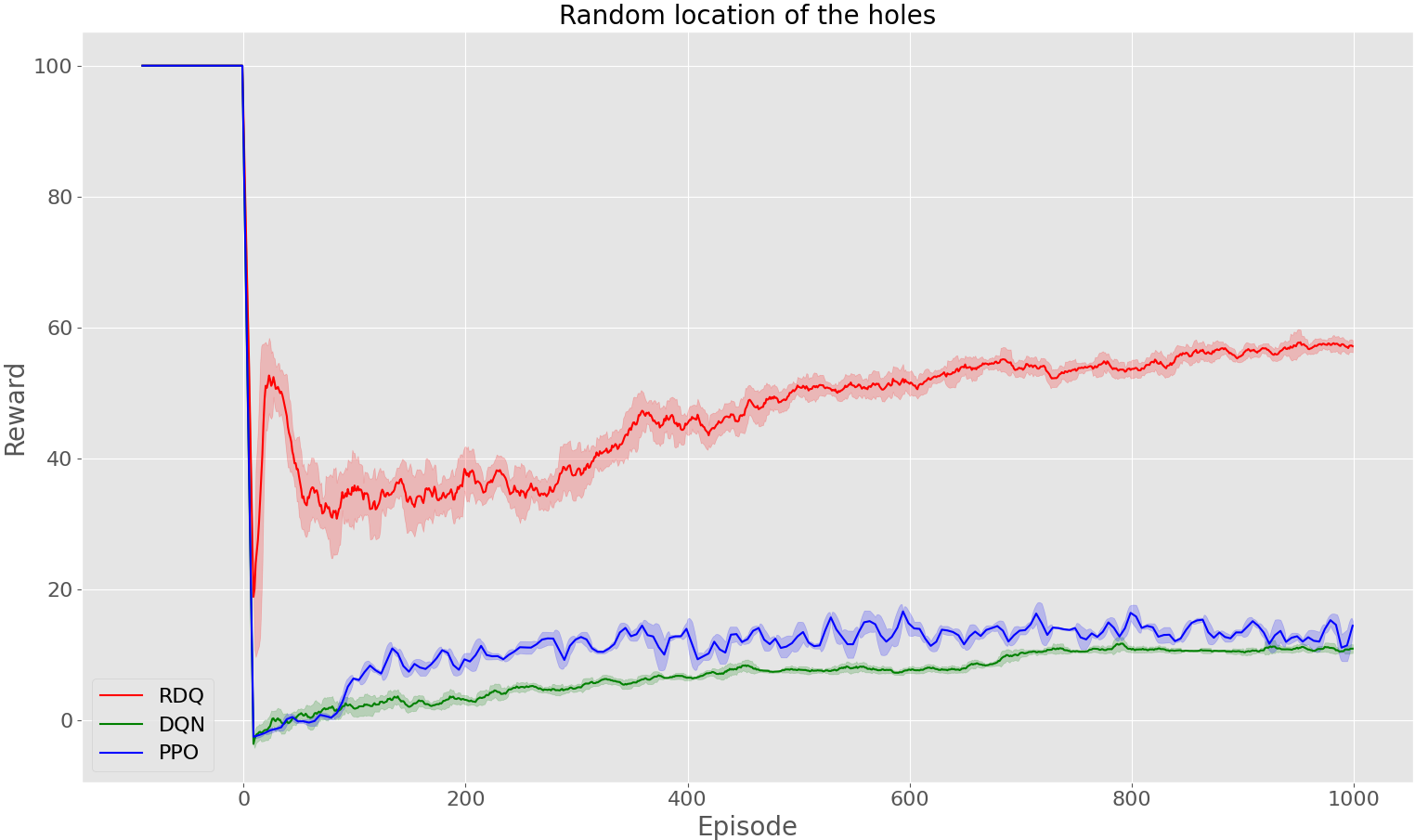}}
\subfloat{\includegraphics[width=.49\linewidth]{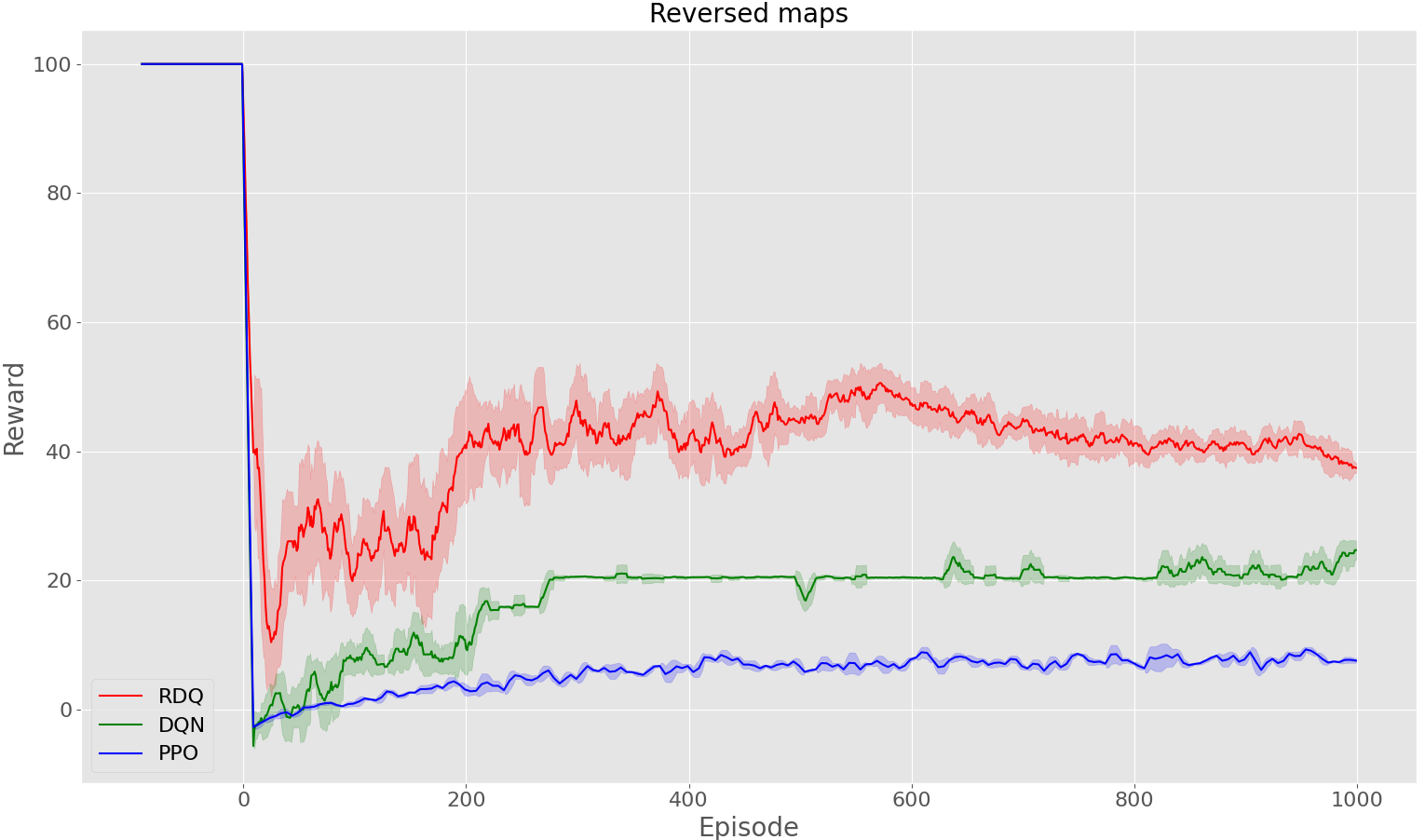}}
\caption{ FrozenLake average results over 100 levels per each novelty. Here negative episodes represent pre-novelty performance. }
\label{fig:res-frozenlake}
\end{figure*}

\paragraph{\textbf{FrozenLake}}
Figure \ref{fig:res-frozenlake} shows the overall average results obtained by DQN, PPO and RDQ agents in 2 types of novelties (i.e. 200 levels with 100 levels per each novelty type). Similarly to Crossroad, all agents were trained to their maximum performance on the base level before encountering the novelties. RDQ agent shows comparable results as in Crossroad, outperforming the DQN and PPO agents in both novelties.

\begin{figure*}
\centering
\includegraphics[width=1.0\linewidth]{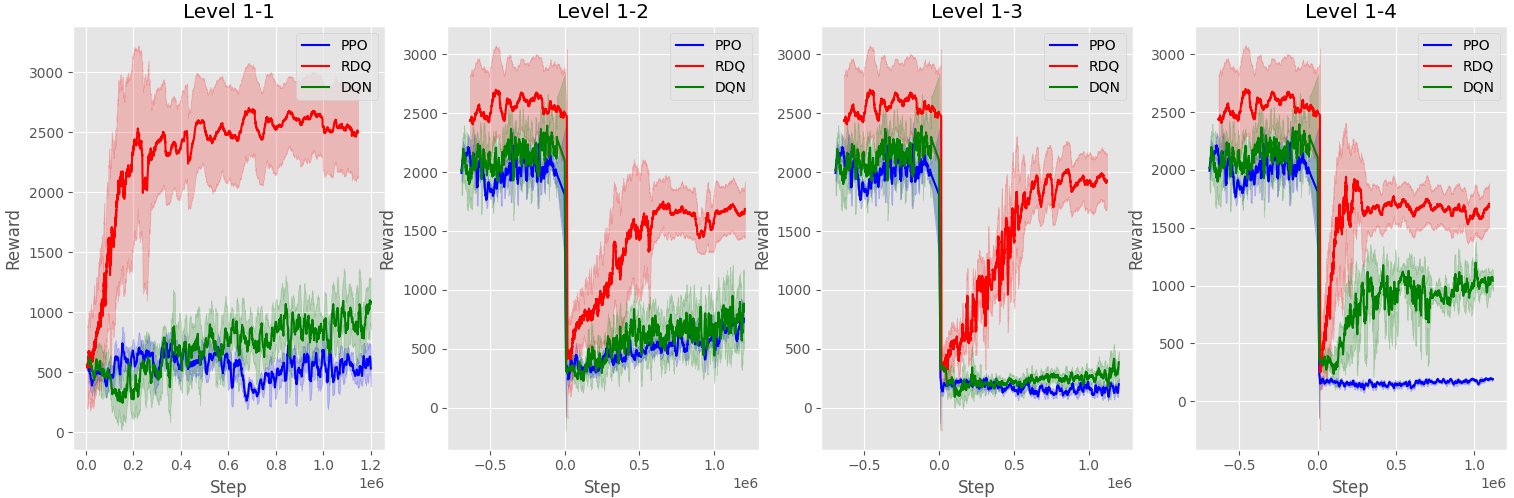}
\caption{ Super Mario Bros average results for each level with pre-novelty and post-novelty performance. Level 1-1 is a base level and has no pre-novelty. }
\label{fig:res-mario}
\end{figure*}

\paragraph{\textbf{Super Mario Bros}}
Figure \ref{fig:res-mario} shows the results of training the RDQ agent on different levels including a base level 1-1. Here we note that due to the dramatic difference between the levels, the RDQ agent shows little resilience to the novelties. We hypothesize that it is due to the completely different observation states (i.e. Figure \ref{fig:mario-levels}) in the levels. Despite that, RDQ shows significantly quicker adaptation to the new levels, highly outperforming both PPO and DQN baseline agents at all levels.

\paragraph{\textbf{Improved Adaptation Speed}}
As mentioned previously, our hypothesis was that by preventing the agent from performing unsafe actions and by teaching it to avoid them, we can improve learning efficiency. Figures \ref{fig:res-crossroad}, \ref{fig:res-frozenlake} and \ref{fig:res-mario} show that this hypothesis stands true in the empirical results. The RDQ agent showed significantly quicker adaptation to all novelties in each of the three domains drastically decreasing needed learning time. 

\paragraph{\textbf{Increased Resilience}}
In addition to being more efficient and explainable, the RDQ agent showed to be more resilient to the novelties in comparison to the baseline agents. Thus for example, in the domains where symbolic rules are highly-transferable (i.e. Crossroad) the agent showed almost no decrease in the overall performance, adjusting its policy in very few steps. However, the agent is still not fully resilient to all types of novelties as seen in Super Mario Bros (Figure \ref{fig:res-mario}). If there is a high degree of novelty (i.e. completely new level) the agent does not have enough knowledge to deal with it. We hypothesize that this drop can be minimized with a larger model trained on the bigger number of levels, but leave it to future work.

\begin{figure}
\centering
\includegraphics[width=.3\linewidth]{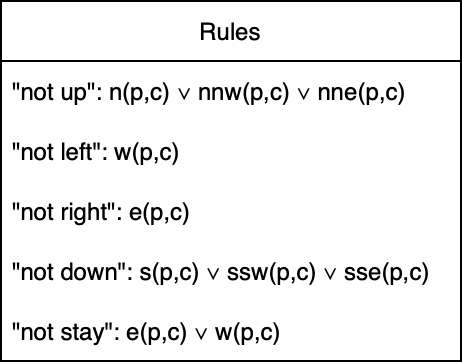}
\caption{ An example of automatically inferred rules for Crossroad. Here $p$ is player, $c$ is car and $n(), w(),...$ are spatial relationships between the objects (north, west, resp.). "Not up" means that action "up" cannot be performed if a relationship is present. Granularity 16 was used for the QSR language. Rules are converted to the human-readable format. }
\label{fig:learnedrule}
\end{figure}

\paragraph{\textbf{Explainability}}
The rule-learning component of the RDQ agent provides human-readable rules that are learned by the agent for each domain (Figure \ref{fig:learnedrule}). Understanding why the agent has made that or another decision is important to ensure that the agent can interact with the real world safely. In addition to being explainable, symbolic rules provide the ability for the human to easily modify them without the need to retrain the agent over a long period of time. All the rules learned by the agent are stored in a single text file that can be modified by the human at any time without interrupting the agent.

\section{Conclusion}
In this work, we proposed a general framework that can be used with deep reinforcement learning agents for a quicker adaptation to the novelties. In our framework, we combined several different AI techniques including inductive logic programming, qualitative spatial representation, knowledge distillation, and safe exploration to significantly improve adaptation speed of the deep reinforcement learning agents. We leveraged and built upon previous work done by the researchers and proposed a new way to incorporate the rules into the learning process as part of the rule-driven Q-learning. We showed one of the possible implementations of the proposed framework as part of the RDQ agent. We empirically demonstrated that the RDQ agent was able to autonomously discover the rules from the negative observations and use them to self-supervise it's learning. Our experiments showed that by using our framework, the RDQ agent outperformed baselines in the tested domains in the adaptation speed and overall resilience to the novelties.

One of the limitations of the RDQ agent is that it can only learn rules to prevent immediate failures and does not consider the consequences of its actions beyond one time-step. Such restriction can be overcome by using more sophisticated rule learning techniques or model-based learning.

Another limitation is that learned rules are disregarded once the novelty is encountered. A better approach would be to partially update the rules that are no longer valid or infer more generic rules, but we leave it to the future work. 

Finally, in this work we focus only on spatial relationships between the objects, however there could be other important relationships between the objects. This limitation can be removed by using and other symbolic language or its combination.

Overall, despite those limitations, RDQ agent was able to outperform baseline agents in all tested domains, providing faster training, improved resilience and efficient adaptation to the novelty.



 \bibliographystyle{elsarticle-num} 






\end{document}